\documentclass[pdflatex, sn-mathphys-num]{sn-jnl}
\usepackage{times}
\usepackage{amsmath,amsfonts,bm,amssymb} 
\usepackage{lineno}
\usepackage{color}
\usepackage{algorithm}
\usepackage{algpseudocode}
\usepackage{xspace}
\usepackage{booktabs}

\newcommand{\alg}{\textsc{SimPEL}\xspace}

\newcommand{\set}[1]{\left\{#1\right\}}
\newcommand{\normaldist}[3]{\mathcal{N}\left(#1 \middle|#2, #3\right)
}

\def\eqref#1{equation~\ref{#1}}
\def\1{\bm{1}}

\newcommand{\norm}[1]{\left\lVert#1\right\rVert}

\def\rvh{{\mathbf{h}}}

\def\vepsilon{{\bm{\epsilon}}}

\def\vmu{{\bm{\mu}}}
\def\vtheta{{\bm{\theta}}}
\def\vphi{{\bm{\phi}}}
\def\va{{\bm{a}}}

\def\vg{{\bm{g}}}
\def\vh{{\bm{h}}}

\def\vs{{\bm{s}}}

\def\vx{{\bm{x}}}
\def\vy{{\bm{y}}}

\def\mI{{\bm{I}}}

\def\mSigma{{\bm{\Sigma}}}

\DeclareMathAlphabet{\mathsfit}{\encodingdefault}{\sfdefault}{m}{sl}
\SetMathAlphabet{\mathsfit}{bold}{\encodingdefault}{\sfdefault}{bx}{n}

\newcommand{\E}{\mathbb{E}}

\newcommand{\R}{\mathbb{R}}

\DeclareMathOperator*{\argmax}{arg\,max}
\DeclareMathOperator*{\argmin}{arg\,min}

\DeclareMathOperator{\divergence}{div}
\DeclareMathOperator{\poly}{poly}

\newcommand{\defeq}{\overset{\text{def}}{=}}

\newcommand{\bX}{\mathbf{X}}

\newcommand{\bx}{\mathbf{x}}
\newcommand{\bs}{\mathbf{s}}

\newcommand{\bU}{\mathbf{U}}

\newcommand{\bh}{\mathbf{h}}
\newcommand{\bg}{\mathbf{g}}

\newcommand{\by}{\mathbf{y}}
\newcommand{\bK}{\mathbf{K}}
\newcommand{\bI}{\mathbf{I}}

\newcommand{\calD}{\mathcal{D}}

\newcommand{\calN}{\mathcal{N}}

\newcommand{\calX}{\mathcal{X}}
\newcommand{\calY}{\mathcal{Y}}

\usepackage{hyperref}
\usepackage[capitalise]{cleveref}
\usepackage{url}
\usepackage{graphicx}
\usepackage{caption}
\usepackage{subcaption}
\usepackage{tikz}
\usepackage{wrapfig}
\usepackage[percent]{overpic}

\title{
Simulation Priors for Data-Efficient Deep Learning}
\begin{document}

% \linenumbers

\author*[1]{\fnm{Lenart} \sur{Treven}}\email{lenart.treven@inf.ethz.ch}
\equalcont{These authors contributed equally to this work.}

\author[1]{\fnm{Bhavya} \sur{Sukhija}}\email{bhavya.sukhija@inf.ethz.ch}
\equalcont{These authors contributed equally to this work.}

\author[1]{\fnm{Jonas} \sur{Rothfuss}}\email{jonas.rothfuss@inf.ethz.ch}
\equalcont{These authors contributed equally to this work.}

\author[1]{\fnm{Stelian} \sur{Coros}}\email{stelian.coros@inf.ethz.ch}
\author[1]{\fnm{Florian} \sur{Dörfler}}\email{dorfler@ethz.ch}
\author[1]{\fnm{Andreas} \sur{Krause}}\email{krausea@ethz.ch}

\affil*[1]{\orgname{ETH Zürich}, \orgaddress{\city{Zürich}, \country{Switzerland}}}

% \affil[2]{\orgdiv{Department}, \orgname{Organization}, \orgaddress{\street{Street}, \city{City}, \postcode{10587}, \state{State}, \country{Country}}}

% \affil[3]{\orgdiv{Department}, \orgname{Organization}, \orgaddress{\street{Street}, \city{City}, \postcode{610101}, \state{State}, \country{Country}}}

\abstract{
How do we enable AI systems to efficiently learn in the real-world?  First-principles models are widely used to simulate natural systems, but often fail to capture real-world complexity due to simplifying assumptions. In contrast, deep learning approaches can estimate complex dynamics with minimal assumptions but require large, representative datasets. 
We propose \alg, a method that efficiently combines first-principles models with data-driven learning by using low-fidelity simulators as priors in Bayesian deep learning. This enables \alg to benefit from simulator knowledge in low-data regimes and leverage deep learning’s flexibility when more data is available, all the while carefully quantifying epistemic uncertainty. We evaluate \alg on diverse systems, including biological, agricultural, and robotic domains, showing superior performance in learning complex dynamics. For decision-making, we demonstrate that \alg bridges the sim-to-real gap in model-based reinforcement learning.
On a high-speed RC car task, \alg learns a highly dynamic parking maneuver involving drifting with substantially less data than state-of-the-art baselines. These results highlight the potential of \alg for data-efficient learning and control in complex real-world environments.
%%% This is too long abstract
% First-principled models are ubiquitously used to simulate natural systems, such as weather, robotics, and biological cells. However, due to their simplifying assumptions, these models do not fully account for the complexity of the real-world. 
% While data-driven approaches can learn system dynamics with minimal assumptions, they require large amounts of data.
% In this work, we propose \alg, an algorithm that efficiently combines first-principled models with data-driven approaches. \alg utilizes
% low-fidelity simulators as priors for Bayesian deep learning. Accordingly, it benefits from the knowledge in the simulator in a low-data regime and the flexibility of deep learning when more data is available.
% We demonstrate the effectiveness of \alg across various applications, including dynamics learning in biological, agricultural, and robotic systems. To demonstrate utility in decision making, we show that \alg successfully bridges the sim-to-real gap between simulation and a high-performance RC car. Using model-based reinforcement learning, we achieve a highly dynamic parking maneuver with drifting, requiring less than half the data compared to state-of-the-art methods.
}


%%\pacs[JEL Classification]{D8, H51}

%%\pacs[MSC Classification]{35A01, 65L10, 65L12, 65L20, 65L70}

\maketitle

% MAIN CONTENT

%%
% \section*{Introduction}
\looseness=-1
For many decades, differential equations have been devised to simulate real-world phenomena such as single-cell
transcriptomics~\citep{dibaeinia2020sergio},  crop growth~\citep{tap2000economics}, options pricing~\citep{black1973pricing}, and robotic systems~\citep{siciliano2016robotics}.
%physical equations of motion have been leveraged to perform highly dynamic and complex tasks in robotics. 
%In many application domains of machine learning, scientists have advanced our understanding of the studied system over decades or even centuries. For instance, in robotics, we have a fairly good grasp of the kinematic and dynamic behavior of our robots. Often such domain knowledge is embedded in some form of differential equations or simulators developed by the respective research community.
However, real-world systems are complex and often do not align with the assumptions made in these first principle simulators, leading to an inherent `sim-to-real' gap. 
%The gap is primarily because domain-specific models/simulators often tend to neglect complex real-world phenomena (e.g., aerodynamics, system latency, elastic deformation, etc.~\cite{widmer2023tuning, hwangbo}) %Accordingly, the 
%In addition, they can only benefit from empirical observations/data through system identification, i.e., to estimate the small number of parameters of the domain-specific model. Compared to modern machine learning models such as neural networks, they unfortunately lack the flexibility to make highly accurate predictions when given sufficient data.

\looseness=-1
Recent advances in machine learning have led to a widespread application of data-driven methods to learn models from data.

\begin{figure}[ht!]
    \centering
    \includegraphics[width=\linewidth]{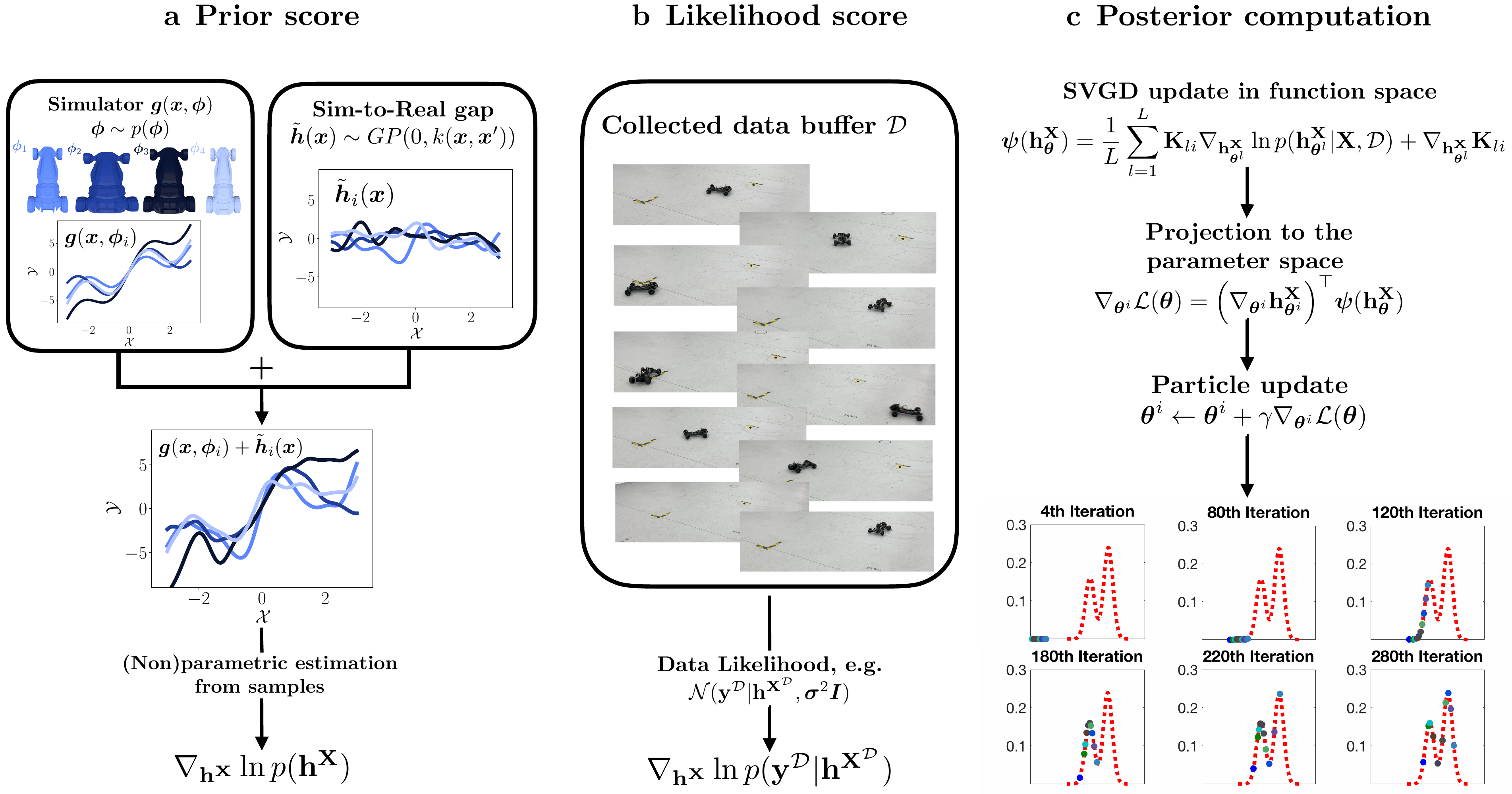}
    \caption{\looseness-1 
(a) \alg estimates the prior score by summing the simulator prior and the sim-to-real GP prior. The score is then estimated either parametrically or non-parametrically from the samples.  
(b) Data is collected and stored in the buffer $\mathcal{D}$. The data likelihood can be modeled with any general likelihood and we model it with a Gaussian, allowing for a closed-form expression of the score.  
(c) Particles iteratively transition from the prior to the posterior distribution. In each iteration, an SVGD update is first performed in function space. This update is then projected from function space to parameter space, followed by a gradient step.  
    }
    \label{fig:panel_0}
\end{figure}

However, these methods often ignore the abundance of knowledge available in simulators and directly learn a model from scratch instead. This is a cause of data inefficiency, since simulators, while imperfect, still capture key system behaviors that would otherwise need to be learned from data. For example, in autonomous cars (\Cref{fig:panel_0}), a bicycle model may not capture nonlinear drifting behaviors, but it still reflects important dynamics like the relationship between throttle, velocity, and steering. Incorporating such knowledge into the learning process can greatly improve sample efficiency.

\looseness=-1
In this work, we explore how to combine the strengths of machine learning with the inductive biases provided by first-principles simulators. We propose a method that retains the flexibility of data-driven models while incorporating domain knowledge encoded in simulators. To achieve this, we build on recent advances in Bayesian inference in function space~\cite{wang2019function, sun2019functional} to impose a functional prior derived from simulation (see \cref{fig:panel_0}), and use it to estimate the prior score from samples. This functional prior acts as a regularizer, encouraging the model to remain close to the simulated dynamics in low-data regimes while allowing it to adapt to real-world dynamics as more data becomes available. Empirically, this leads to a significant improvement in sample efficiency.
% In this paper, we discuss how we can harness the benefits of both machine learning and first principles simulators. We propose a method that enjoys the flexibility of data-driven approaches while also retaining the domain knowledge embedded in simulators.
% Moreover, we leverage recent advances in Bayesian inference in functional space~\cite{wang2019function, sun2019functional} to impose a functional prior derived from the simulation (see \cref{fig:panel_0}) and use it to esitmatee the prior score from the samples. The prior regularizes the model to behave similarly to the simulation in a low-data regime and fit the real-world dynamics when data is available. Empirically, this results in a substantial sample-efficiency gain. 

\looseness=-1
%In summary, our main contributions are; (\emph{i}) we propose \alg, a simple and tractable approach for incorporating available simulators in the training of Bayesian Neural Networks (BNNs), (\emph{ii}) we ablate different choices of score estimators for the simulation-based functional prior, (\emph{iii}) we evaluate the performance of \alg across several domains ranging from biology, agriculture to robotics. Furthermore (\emph{iv}) we combine \alg with a model-based reinforcement learning algorithm and evaluate it on a highly dynamic RC Car (cf.,~\cref{fig:panel_0}). We
%show that \alg results in considerably better performance in the offline-RL setting and significantly
%faster convergence for the online RL case.
In summary, our main contributions are; (\emph{i}) we propose \alg, a simple and tractable approach for incorporating available simulators in the training of Bayesian Neural Networks (BNNs), (\emph{ii}) we evaluate the performance of \alg across several domains ranging from biology, agriculture to robotics, and (\emph{iii}) we combine \alg with a model-based reinforcement learning algorithm and evaluate it on a highly dynamic RC Car (cf.,~\cref{fig:panel_rl}). 
We show that \alg results in considerably better performance in the offline-RL setting and significantly faster convergence for the online RL case. Finally, (\emph{iv}) we propose and ablate several variants of \alg based on different score estimation techniques for the simulation-based functional prior.
\subsection*{Incorporating Simulators as Functional Priors}
\label{subsection: Incorporating simulators as functional priors}
\looseness = -1 %In this section, we present \alg. 
We illustrate the main idea behind \alg on 
%consider 
a simple pendulum system.
%as a motivating example to illustrate the main idea behind \alg.   
The pendulum's state $\bs = [\varphi, \dot{\varphi}]$ is represented through the angle $\varphi$ and angular velocity $\dot{\varphi}$. The following ODE is commonly used to describe the pendulum dynamics
\begin{align} \label{eq:pendulum_model}
    \ddot{\varphi} = \frac{mgl \sin(\varphi) + C_m u }{I} ~.
\end{align}
\looseness =-1  Here $m, l$ and $I$ are the mass, length, and moment of inertia, respectively. %is the mass, $l$ the length, and $I$ the moment of inertia of the pendulum. 
% The control input to the system is the torque $\tau = C_m u$.
The motor at the rotational joint of the pendulum applies the torque $\tau = C_m u$ proportional to the input $u$. 

%There are two key sources of uncertainty/error in the dynamics of the real pendulum.
%First, the exact parameters $\vphi$ (here, $\vphi=[m, l, C_m, I]$) of the domain-specific model are unknown. 

To accurately predict the behavior of a real-world pendulum system, we have to deal with two sources of uncertainty/error. First, the exact parameters $\vphi$ (here, $\vphi=[m, l, C_m, I]$) of the domain-specific model are unknown. 
However, we can typically narrow down each parameter's value to a plausible range. We capture this via a prior distribution $p(\vphi)$ over the model parameters, akin to the practice of domain randomization in robotics.  %The process of randomly sampling a parameter set $\vphi \sim p(\vphi)$ and then integrating/solving the ODE in \Cref{eq:pendulum_model} with the corresponding parameters gives random functions. This allows us to implicitly construct a stochastic process of functions that reflect our simplified pendulum model. We call this the {\em domain-model process}.
The second source of uncertainty is various physical phenomena such as aerodynamic drag, friction, and motor dynamics that are not captured in (\ref{eq:pendulum_model}). This results in a systematic sim-to-real gap. %only gives a non-zero probability to transition functions that reflect linear motor torque output, no friction and no drag. 

\looseness=-1
In \alg, we model the problem of learning the system dynamics as a general regression problem with a dataset $\mathcal{D}=(\mathbf{X}^{\mathcal{D}},\mathbf{y}^{\mathcal{D}})$, comprised of $m$ noisy evaluations $\by_{j}=\bh^*\left(\bx_{j}\right) + \vepsilon_j$ of an unknown function $\bh^*: \calX \mapsto \calY$ with $\calX \subseteq \R^{d_x}$ and $\calY \subseteq \R^{d_y}$. 
The training inputs are denoted by $\mathbf{X}^{\mathcal{D}} = \left\{ \bx_j \right\}_{j=1}^m$ and corresponding function values by $\mathbf{y}^{\mathcal{D}} = \left\{ \by_j \right\}_{j=1}^m$. For the pendulum example, the input consists of the state and input $\vx = [\vs, u]$, the output is the next state, i.e., $\vy = \vs'$, and $\bh^*$ is the system dynamics.

We incoproate the knowledge embedded in low-fidelity simulators such as \cref{eq:pendulum_model} by
 constructing a simulation prior $p(\bh)$ in
the space of regression functions $\bh: \calX \mapsto \calY$. Moreover, we randomly sample parameters $\vphi \sim p(\vphi)$ and then integrate the ODE in \Cref{eq:pendulum_model} with the corresponding parameters. This constructs a stochastic process of functions that reflect our simplified pendulum model, which we can use as a functional prior for learning the true pendulum dynamics.
We factorize the prior over the output dimensions, i.e., $p(\bh) = \prod_{i=1}^{d_y} p(h_i)$. This allows us to treat each $h_i: \bX \mapsto \R$ as an independent scalar-valued function.

However, our prior based on the simulation model does not account for the sim-to-real gap.  Next, we discuss in further detail how we represent the simulation prior and incorporate the sim-to-real gap.

%\subsection*{Incorporating simulators as functional priors}

\paragraph{Domain-Model Process} \looseness=-1 The first component of the implicit prior is the domain-specific (low-fidelity) simulation model $\bg(\bx, \vphi)$, which maps each input-parameter tuple $(\bx, \vphi)$ to $\calY$. Crucially, we do not require an analytical expression of $\bg: \calX \times \Phi \mapsto \calY$. It is sufficient to have query access, i.e., the output vector of $\bg(\bx, \vphi)$ can be the result of a numerical simulation (e.g., ODE, SDE solver, rigid body simulator, agent-based simulation etc.). 

\paragraph{Sim-to-Real Gap Prior} \looseness=-1 
The second component is the \emph{sim-to-real gap process}. 
To model this, we employ a Gaussian process (GP) $ p(\tilde{h}_{i}) $ for each output dimension $ i = 1, \dots, d_\vy $. The role of this GP is to capture the discrepancy between the domain model $ g_i(\bx, \vphi) $ and the true target system $\bh^*(\bx)$. We choose a GP for its generality, simplicity, and interpretability. Specifically, we use a GP with zero mean and an isotropic kernel of the form $k(\bx, \bx') = \kappa^2 \rho\left(\norm{\bx - \bx'}/\ell\right)$,
where $ \kappa^2 $ is the variance, $ \ell $ is the lengthscale, and $ \rho $ is a positive-definite correlation function (e.g., squared exponential or Matérn). The hyperparameters $ \ell $ and $ \kappa^2 $ encode prior beliefs about the nature of the sim-to-real gap. A small variance $ \kappa^2 $ suggests that the gap is expected to be small, while a large lengthscale $ \ell $ indicates that discrepancies from the domain model are smooth and systematic rather than highly localized.

% The second component is the sim-to-real gap process. We employ a GP $p(\tilde{h}_{i})$ per output dimension $i=1, ..., d_y$ for this purpose. The GP aims to model the gap between the domain model $g_i(\bx, \vphi)$ and the actual target function $f_i(\bx)$. We use a GP model for this purpose since it is very general, simple, and interpretable. Moreover, we use a GP with zero mean and an isotropic kernel $k(\bx, \bx') = \kappa^2 \rho(\norm{\bx-\bx'} / \ell)$ with variance $\kappa^2$ and lengthscale $\ell$.
% The lengthscale $l$ and $\kappa^2$ influence our belief on the sim-to-real gap.
% A small $\kappa^2$ implies that the sim-to-real gap is small
% and a large length scale conveys that deviations from the domain model are systematic rather than local.
 
\looseness=-1
Given a set of measurement points $\bX$, the marginal distributions of the sim-to-real prior follow a multivariate normal distribution $p(\tilde{\bh}^{\bX}_{i}) = \calN(\tilde{\bh}^{\bX}_i | \mathbf{0}, \bK )$ with kernel matrix $\bK = [k(\bx_i, \bx_j)]_{i,j}$.

\paragraph{Combining the Stochastic Processes}
Stochastic processes can be viewed as infinite-dimensional random vectors, and therefore, representing $p(\vh)$ is intractable. Instead, we sample a set of measurement points from the domain $\mathcal{X}$, i.e., $\bX \sim \bm{\zeta}$, where $\bm{\zeta}$ is a measurement distribution, e.g., $\operatorname{Unif}(\calX)$.
Given $\bX$, we then independently sample (conditional) random vectors from both processes and add them together.
\begin{align} \label{eq:combined_sampling}
\begin{split}
\rvh_i^{\bX} = \hspace{2pt} & [g_i(\bx_1, \vphi), ..., g_i(\bx_k, \vphi)]^\top + \tilde{\bh}^{\bX}_i \\ & \text{with}~~ \vphi \sim p(\vphi) , ~  \tilde{\bh}_i^{\bX} \sim \calN(\tilde{\bh}_i^{\bX}| \mathbf{0}, \bK ).
\end{split}
\end{align}
\looseness -1
The resulting stochastic process prior $p(\bh)$ is defined implicitly through the marginal distributions that are implied by (\ref{eq:combined_sampling}).
We use the estimated stochastic process $p(\bh^{\bX})$ as a prior to train our model. 

\paragraph{Posterior update}
Uncertainty quantification plays a crucial role in many machine learning applications, e.g., exploration in online reinforcement learning~\citep{chua2018deep, curi2020efficient, sukhija2024optimistic, sukhija2025optimism} or active learning~\citep{settles2009active, krause2008near, balcan2010true, hubotter2024informationbased}, and safety certification for safe learning applications~\citep{pmlr-v37-sui15, berkenkamp2021, racecarSafeopt, siemensSafeOpt, sukhija2023gosafeopt, widmer2023tuning}. Uncertainty-aware models, such as Bayesian neural networks (BNNs), estimate the underlying function $\bh^*$ but also quantify the uncertainty in their estimates. Given the importance of uncertainty quantification, in this work, we focus on Bayesian deep learning models.
%capturing the confidence/epistemic uncertainty of the learned model is very important. Accordingly, we train a Bayesian neural network model (BNN) model of $\bh^*$.  
%We leverage the functional prior $p(\bh^{\bX})$ to
 %train a Bayesian neural network model (BNN) model.
 Moreover, we employ a parameterized model $\bh_\vtheta: \calX \rightarrow \calY$ with weights $\vtheta \in \Theta$. 
However, instead of imposing a prior on the parameters $\vtheta$, we use 
the functional prior $p(\bh^{\bX})$. This results in the posterior~\citep{wang2019function, sun2019functional} 
\begin{equation}
   p(\bh |\mathcal{D}) \propto p(\mathbf{y}^{\mathcal{D}} | \mathbf{X}^{\mathcal{D}},\bh) p(\bh^{\bX}).
   \label{eq: functional posterior}
\end{equation}
\looseness=-1
Determining $p(\bh |\mathcal{D})$ is intractable for most distributions and models. To this end, approximate Bayesian inference techniques, such as particle-based methods, are commonly used to estimate the posterior. Furthermore, note that while we focus on Bayesian deep models,  the right side of \cref{eq: functional posterior} could already be used to obtain a maximum a posteriori (MAP) estimate for $\vtheta$. 
 %We use approximate Bayesian inference techniques, in particular, particle based methods to approximate the posterior $p(\bh |\mathcal{D})$. 
 
Additional details on how to incorporate the prior and the model training/Bayesian inference 
are provided in the \textbf{Method} section and the algorithm is summarized in \cref{alg:sim_transfer_alg_short}.

%Here, $p(\bh)$ is a {\em stochastic process} prior, e.g., Gaussian process (GP)~\citep{rasmussen2003gaussian}. 

%with the FSVGD from \cref{eq:fsvgd_updates}. The score for the simulation-based stochastic process is given by
% \begin{equation} \label{eq:score_sim}
%        \nabla_{\rvh^\bX} \ln p(\rvh^\bX |\mathbf{X},  \mathcal{D})  = \nabla_{\rvh^\bX} \ln p(\mathbf{y}^{\mathcal{D}} | \rvh^{\bX^{\mathcal{D}}}) +  \sum^{d_y}_{i=1}\nabla_{\rvh_i^\bX}\ln p(\rvh_i^{\bX}) ~
% \end{equation}
%  In the following, we discuss different approaches for obtaining the score.

\begin{algorithm}[t]
\caption{\strut \alg}\label{alg:sim_transfer_alg_short}
\hspace*{\algorithmicindent} \textbf{Input:} Measurement distribution
$\bm{\zeta}$, Simulation prior $\bg$, Parameter distribution $p(\vphi)$, GP 
\hspace*{\algorithmicindent} \hspace{2.7em} $p(\tilde{\bh})$, Data $\calD$, BNN particles $\{\vtheta_i\}^{L}_{i=1}$
\begin{algorithmic}[1]
\State Obtain measurement set $\bX$ using measurment distribution $\bm{\zeta}$.
\State Sample simulation prior function values $\{\bh^{\bX}_{i, j} \sim \bh^{\bX}_{i} \}^{d_y}_{i=1}$ for $j \in \{1, \dots, N\}$ using simulator prior $\bg$, parameter distribution $p(\vphi)$, and GP $p(\tilde{\bh})$ as in \eqref{eq:combined_sampling}.
%\State Approximate the prior score term $\nabla_{\rvh_{\vtheta^l}^\bX} \ln p(\rvh_{\vtheta^l}^\bX)$ for $l \in \set{1, \ldots, L}$ with samples from the prior $\{\bh^{\bX}_{i, j}\}^{d_y}_{i=0}$ for $j \in \{1, \dots, N\}$ using techniques from \Cref{subsection: Estimating the Stochastic Process Prior Score}.
\State Use samples $\{\bh^{\bX}_{i, j} \}_{i=1, j=1}^{d_y, N}$ to estimate the prior score $\nabla_{\rvh^\bX} \ln p(\rvh^{\bX})$ and update the BNN particles $\{\vtheta_i\}^{L}_{i=1}$ as in \eqref{eq:fsvgd_updates}.
\end{algorithmic}
\end{algorithm}

\section*{Related Work}
\label{section: Related Work}

\paragraph{Neural Networks}
\looseness=-1
Neural networks are commonly used for learning general regression tasks from data~\cite{hwangbo, narendraNNSysID, SJOBERG1994359, nagabandi2018, bernSoftRobotControl2020, sukhija2023gradient, krizhevsky2012imagenet} due to their high expressiveness. 
However, in several downstream applications, such as classifying rare diseases from medical images, controlling the robot, and out-of-distribution generalization, using pure NN models is challenging because in the downstream tasks we exploit (``overfit'' to) inaccuracies of the learned model.
%This leads to suboptimal performances ~\cite{chua2018deep}.
BNNs do not suffer from the same pitfall~\cite{chua2018deep, ngartera2024application}. As a result, recently, BNNs have been widely used for learning in applications such as robotics~\cite{chua2018deep, curi2020efficient, sekar2020planning, treven2024ocorl,
sukhija2024optimistic,rothfuss2023hallucinated}. 
% However, during control, planning with NN models is challenging because most planners exploit (``overfit'' to) inaccuracies of the learned model. This leads to suboptimal performance on the real system~\cite{chua2018deep}.
% BNNs do not suffer from the same pitfall~\cite{chua2018deep}. As a result, recently, BNNs have been widely used for dynamics learning~\cite{chua2018deep, curi2020efficient, sekar2020planning, treven2024ocorl,
% sukhija2024optimistic,rothfuss2023hallucinated}. 
\paragraph{Bayesian Deep Learning}
Unlike learning a single NN, Bayesian Neural Nets (BNNs) maintain a distribution over the NN parameters. Moreover, using a known prior distribution over the NN parameters, given the training data, BNNs infer the posterior distribution. Exact Bayesian inference is computationally intractable for NNs, and approximations
% ~\cite{mackay1992bayesian, neal2012bayesian, graves2011practical,blundell2015weight, gal2016dropout, zellers2018swag, goan2020bayesian} are generally used.
such as Laplace~\cite{mackay1992bayesian}, Markov Chain Monte Carlo (MCMC)~\cite{neal2012bayesian}, or variational inference~\cite{graves2011practical, blundell2015weight} are generally used (cf.,~\cite{goan2020bayesian} for a detailed survey). Practically, BNNs are computationally expensive to train and inexpensive alternatives such as  
MC dropout~\cite{gal2016dropout} and SWAG~\cite{zellers2018swag} give overconfident uncertainty estimates~\cite{lakshminarayanan2017simple}. 
In this work, we focus on particle-based BNNs due to their computational tractability and reliable uncertainty estimates ~\cite{wang2019function, lakshminarayanan2017simple, Liu2016} (cf.,~\cite{d2021repulsive} for more details). %Particle-based techniques are computationally tractable, simple to implement and give good uncertainty estimates~\cite{d2021repulsive}. 
Classical BNNs impose a generic standard normal prior on the NN parameters. 
However, such priors tend to form straight horizontal lines \citep{tran2022all}, an effect that arises from increasing the depth of the model \citep{neal2012bayesian,duvenaud2014avoiding,matthews2018gaussian}, and makes the priors not capture the functions modeled well in practice.
On top, priors on weights do not have an intuitive interpretation in terms of the function's output behavior, making it difficult to incorporate domain-specific knowledge directly into the model via weight priors \citep{fortuin2022priors}. 
\citet{sun2019functional} replaces prior over weights with prior over functions for BNN inference.
% and shows the possibility and effectiveness of having priors with rich structures, such as Gaussian processes.
Several works show that priors can also be implicitly incorporated into BNN learning \cite{shi2018spectral,zhou2020nonparametric}.
\alg leverages existing simulators to impose an informed implicit prior and adds additive GP to cover for the gap between the simulator and the modeled data-generating process. This boosts sample efficiency when learning BNNs from data and also enables bridging the `sim-to-real' gap.

% Nearly all BNNs impose a generic standard normal prior on the NN parameters. On the contrary, we leverage existing simulators to impose an informed prior. This boosts sample efficiency when learning robot dynamics from data and also enables bridging the `sim-to-real' gap (cf.,~\cref{section: Experiments}).

\paragraph{Bridging the Sim-to-Real Gap}
Domain randomization, meta-learning, and system identification are typically used to bridge the sim-to-real gap in biology, robotics and RL (cf.,~\cite{clavera2018model, zhao2020sim, rothfuss2021meta, tobin2017domain}). Domain randomization based approaches \cite{tobin2017domain,yue2019domain,mehta2020active,chen2021understanding} assume that the real world is captured well within the randomized simulator. 
This is not the case for many problems, such as those we consider in this work. Meta-learning methods~\cite{duan2016rl, finn2017model, finn2018probabilistic, bhardwaj2023data} also require that the meta-training and test tasks are i.i.d. samples from a task distribution, and thereby representative. Furthermore, they generally require a plethora of data for meta-training, which is typically also computationally expensive.
Instead, \alg does not require any meta training and solely leverages query access to low-fidelity simulators, which are based on first-principles physics models and are often available in robotics and biology \cite{taubes2008modeling,hutter2018robot}. 
We showcase the benefits of \alg on our hardware experiments, where we significantly outperform existing model-based learning approaches in data-scarce regimes.
% In our hardware experiment, we solely leverage real-world data and operate in the data-scarce regime. To this end, we consider model-based learning approaches due to their sample efficiency.
Typical model-based approaches, perform system identification~\cite{ljungSystemID, li2013terradynamics, moeckel2013gait, tan2016simulation, zhu2017model, gofetch} to fit the simulation parameters on real-world data. However, if the simulation model cannot capture the real-world dynamics well, this results in suboptimal performance (cf., \textbf{Results} section). 
Most closely related to our work are approaches such as~\cite{hwangbo, pastor2013learning,  ha2015reducing, HUANG20178969, schperberg2023real}. They use the simulator model and fit an additional model, e.g., NN or Gaussian process, on top to close the sim-to-real gap \cite{li2021grey}. We refer to this approach as \textsc{GreyBox}. In \Cref{fig:1d_sinusoids} we show that our method results in better uncertainty estimates than the \textsc{GreyBox} approach.

\section*{Results}
%\label{section: Experiments}

In this section, we empirically evaluate the benefits of incorporating first-principles priors into learning algorithms.\footnote{The code to reproduce all our experiments is available on \url{https://github.com/lasgroup/simulation_transfer}.}
We begin by assessing how \alg's posterior mean and variance predictions compare to those of state-of-the-art BNN inference methods.
We then apply \alg to the problem of learning complex dynamical models across diverse domains—including robotics, biology, and single-cell dynamics—where sample collection is particularly costly yet low-fidelity models are available. In such data-scarce settings, it is crucial to design algorithms that not only yield accurate mean predictions but also provide reliable uncertainty estimates from limited data.
Finally, we examine whether \alg can improve the sample efficiency of both offline and online reinforcement learning (RL). Since data collection in these RL settings often involves human interventions, enhancing sample efficiency can substantially accelerate learning and reduce associated costs.

\subsection*{Does \alg obtain an accurate and precise Posterior?}
\label{subsection: Illustrative experiment on sinusoids}

\begin{figure}[ht!]
    \centering
    \includegraphics[width=\linewidth]{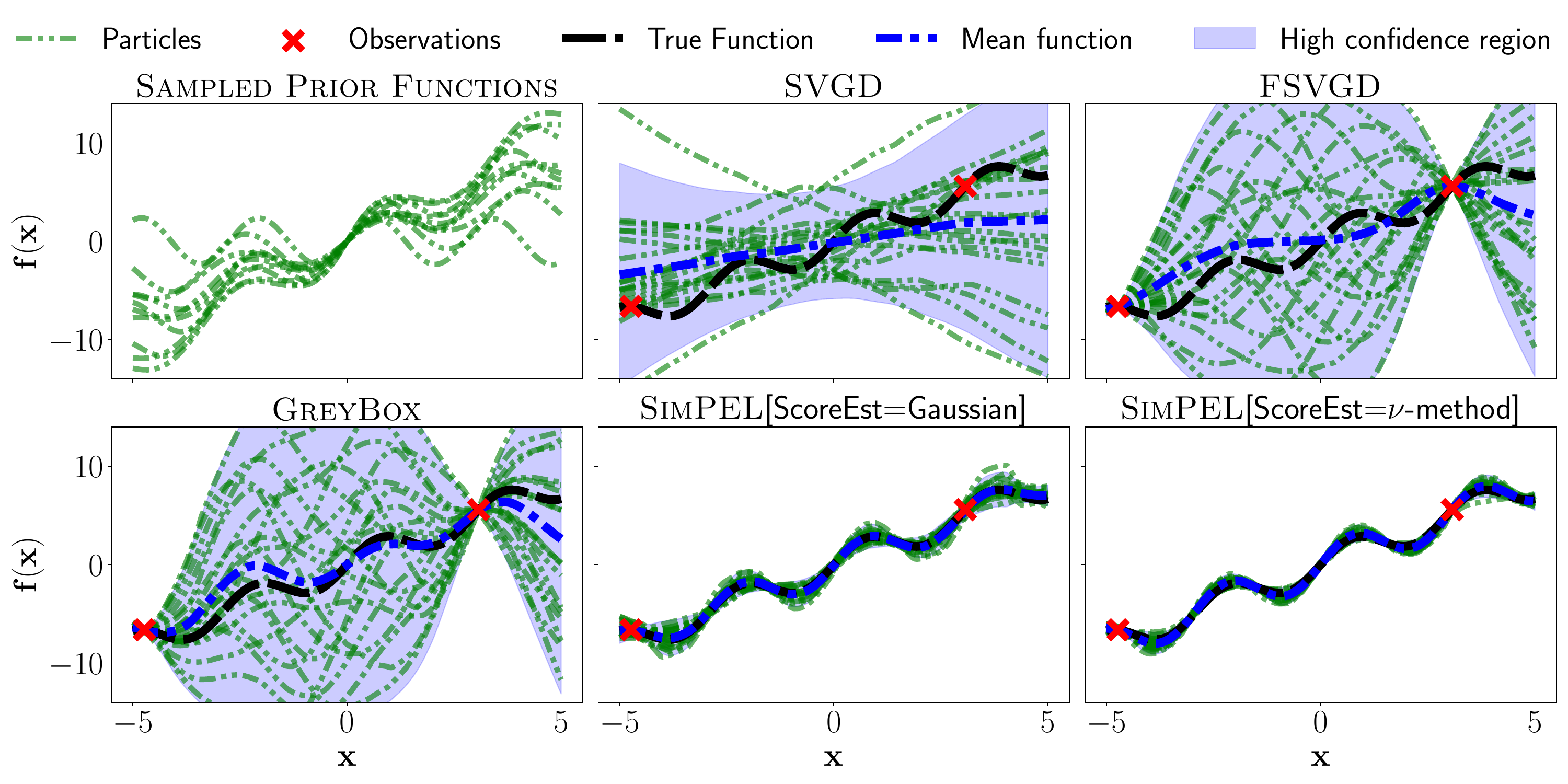}
    \caption{BNN posteriors trained on two data points for a one-dimensional sinusoidal function. 
    \textsc{SVGD} with Gaussian prior on the neural net weights struggles with fitting the mean already on the observation points. \textsc{FSVGD} with Gaussian process prior accurately predicts the function value around the observations, however, on the area without data, the mean is not accurate and uncertainties are large.  
    \textsc{GreyBox} finds accurate mean everywhere but has large uncertainty estimates on the area without data. At the same time \alg, applied with 2 prior score estimation techniques (ScoreEst stands for prior score estimation technique, c.f. \textbf{Method} section), obtains both accurate mean and precise uncertainty estimates.}
    \label{fig:1d_sinusoids}
\end{figure}

We illustrate the performance of \alg on a simple one dimensional example in \cref{fig:1d_sinusoids}. The simulation prior in this example is sinusoids with different amplitudes, frequencies, and slopes. We compare \alg with \textsc{SVGD} \citep{Liu2016}, \textsc{FSVGD} \citep{sun2019functional} and \textsc{GreyBox}, where we first fit the simulation model to the data and then correct for the misspecifications with \textsc{FSVGD}. 
From the \cref{fig:1d_sinusoids} we conclude that \alg fits the mean of the underlying function the best across all our baselines, while also having much more precise uncertainty estimates due to the simulation prior. Although \textsc{GreyBox} achieves a comparable fit to the mean, its uncertainty predictions are overestimated and resemble those of \textsc{FSVGD}. This discrepancy arises because \textsc{GreyBox} relies on low-fidelity models for mean estimation but does not leverage them effectively for uncertainty quantification.

\subsection*{Does \alg improve System Identification?}
\label{subsection: Application to System Identification}
\begin{figure}[th]
    \centering
    \includegraphics[width=\linewidth]{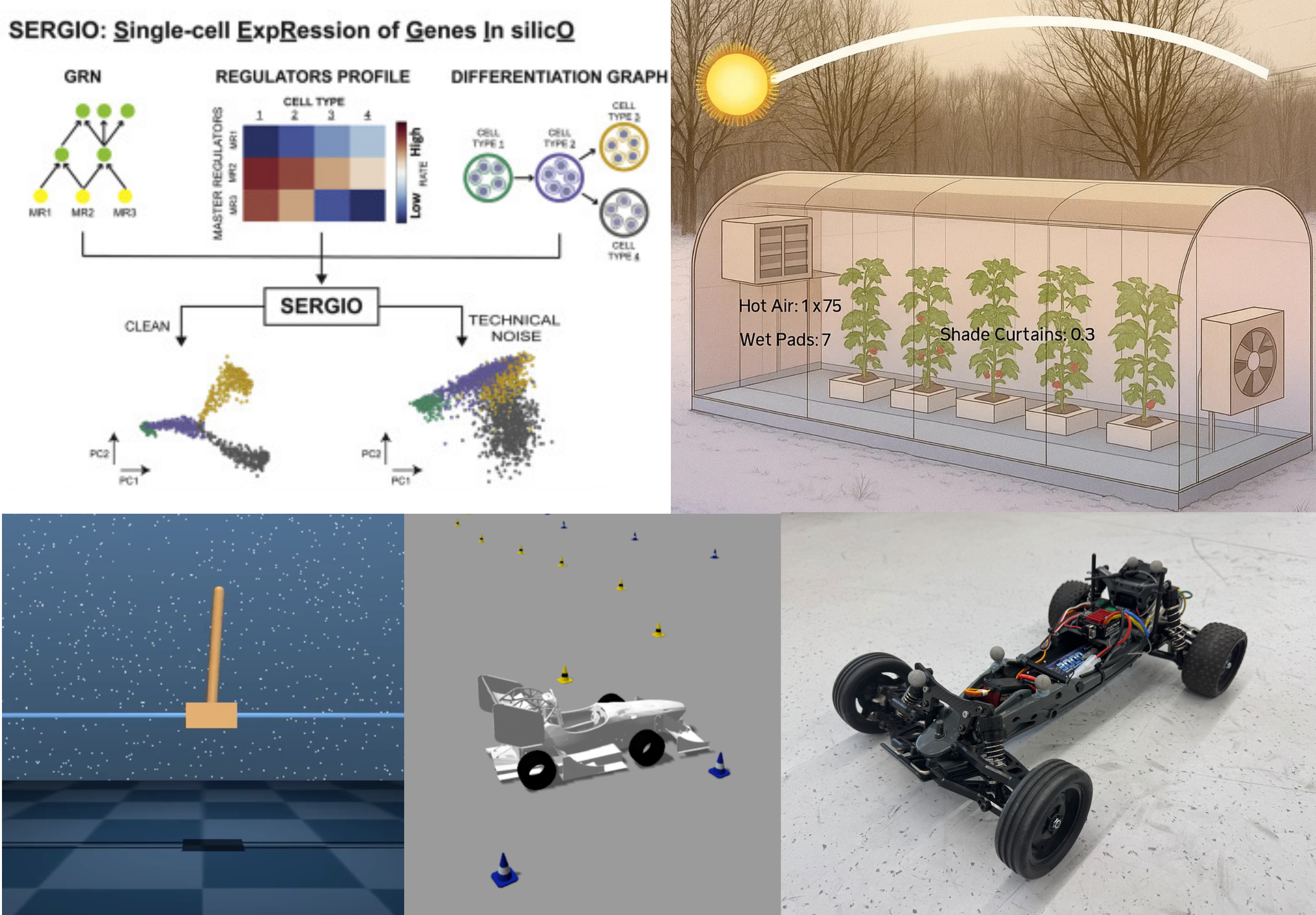}
    \caption{\looseness-1 
We consider environments from three domains: biology, agriculture, and robotics. The top left shows a biological environment with single-cell expression \citep{dibaeinia2020sergio}. The top right represents crop growth, specifically greenhouse climate control \citep{fitz2010dynamic}. The bottom row features robotic environments: a pendulum \cite{tunyasuvunakool2020dm_control} (left), a racecar \citep{kabzan2020amz} (middle), and an RC car (right).
    }
    \label{fig:considered systems}
\end{figure}

We apply \alg for system identification on four different dynamical systems; (\emph{i}) pendulum, (\emph{ii}) single-cell expression, (\emph{iii}) greenhouse climate control, and (\emph{iv}) race car (cf. \cref{fig:considered systems}). For the latter, we also consider data from an RC car hardware platform. All of these systems exhibit complex dynamics and have first-principles based models~\citep{dibaeinia2020sergio, tap2000economics, kabzan2020amz}.

We first evaluate whether incorporating low-fidelity simulators as priors can improve the model fit of high-fidelity simulators for systems \emph{(i)–(iv)}. We then assess whether a similar approach improves model fit for the RC car hardware system using its low-fidelity simulator as a prior.

\paragraph{Low-fidelity to High-fidelity in Simulation}
\begin{figure}[ht!]
    \centering
      \begin{overpic}[width=\textwidth]{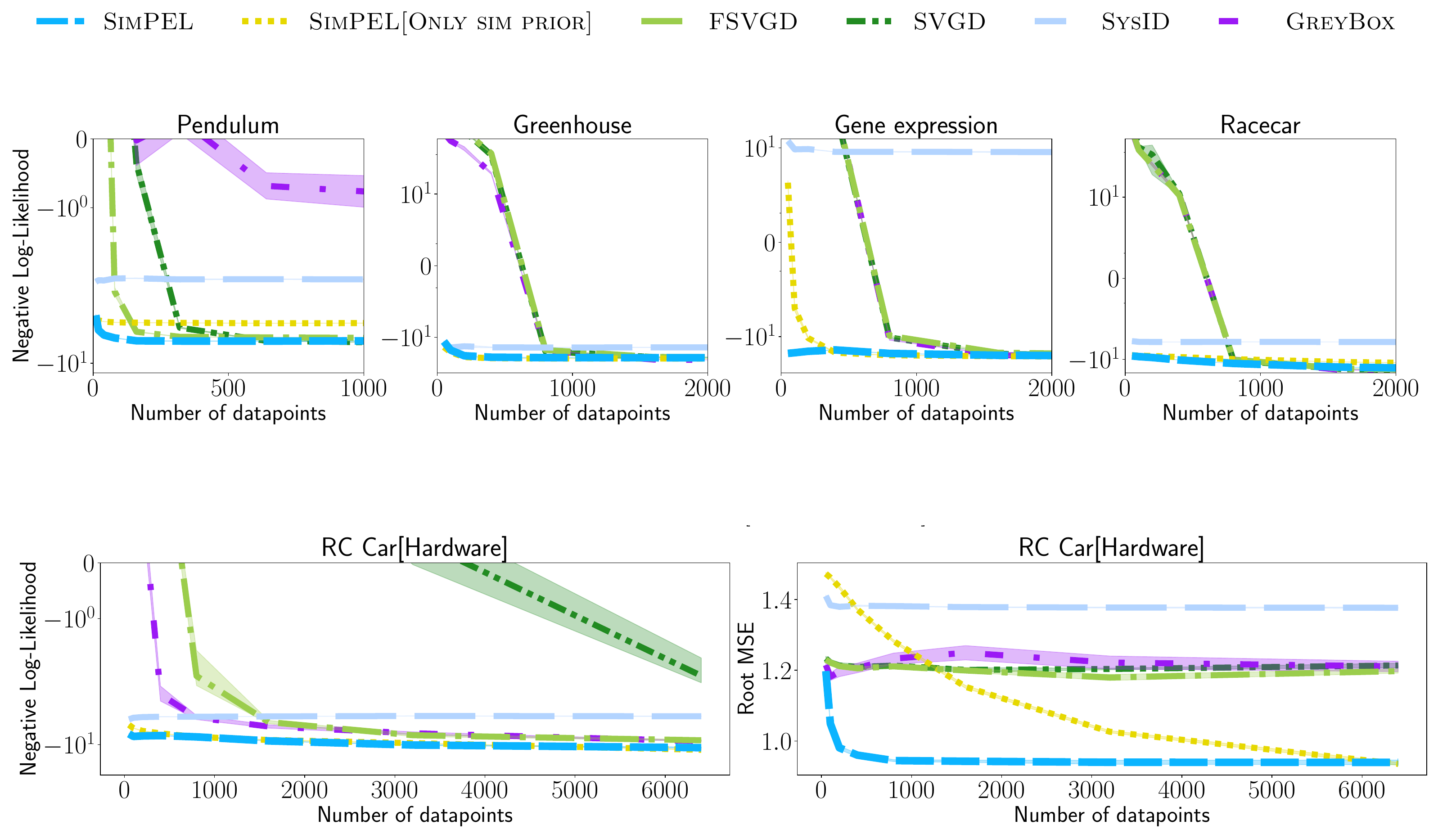}
    \put(0,51){\small \textbf{a) Regression on simulators}}
    \put(0,23){\small \textbf{b) Regression on hardware}}
    \put(0,27){\line(1,0){100}}
  \end{overpic}
    \caption{\looseness-1 
    (a)
    On all the systems \alg achieves low Negative Log-Likelihood already in the low data regime and performs on par with the best from the rest also in high data regime. In Sergio, Greenhouse and Racecar environments, \textsc{SimFSVGD[Only sim prior]} also achieves the performance of \alg, which indicates that the prior over $\vphi$ is rich enough so that it can compensate the missing additive GP term. That does not happen in the Pendulum environment, showcasing the importance of having a GP to model the gap between the models.
    (b)
    Also on the hardware dynamics, \alg achieves low Negative Log-Likelihood score already in low data regime and performs on par with best of the rest in the high data regime. \alg performs the best also with respect to the Root MSE score, with only \textsc{SimSVGD[Only sim prior]} matching its performance in the high data regime.}
    \label{fig:panel_regression}
\end{figure}
For all the systems, we consider simulators of two different complexities. We call the simulator that models only the basic dynamics (in Pendulum, for example, just the gravity) low fidelity simulator, and the simulator that models also more complex dynamics (e.g., friction and air resistance for the pendulum) high fidelity simulator. For the prior stochastic process we take the combination of the low fidelity simulator $\vg(\bx, \vphi)$, where $\vphi$ are unknown parameters of the system (mass and length for the pendulum), and the additive GP to model the gap between the low and high fidelity simulators. 
For the cell expression task, we consider Sergio from \citet{dibaeinia2020sergio} and for the Greenhouse task the simulator from \citet{tap2000economics}.

We learn the dynamics of the high fidelity simulator and study how \alg performs in comparison to other methods. The performance metric for our experiments is the Negative Log-Likelihood and we train the algorithms on train sets of increasing size and have a separate test set on which we evaluate the Negative Log-Likelihood of the trained posterior model. We present the results in \Cref{fig:panel_regression}. We compare \alg with \textsc{FSVGD},  \textsc{SVGD}, and \textsc{GreyBox}. Furthermore. we also consider a version of the algorithm, \textsc{SimFSVGD[Only sim prior]}, which only uses the simulator dynamics for the prior, i.e., no additive GP. Lastly, we also have a \textsc{SysID} baseline where we fit the parameters of the low-fidelity model to the data. 
Across all systems, \alg achieves the best performance. In particular, due to the sim prior and expressive nature of the NN model, \alg does well in the low-data and high data regimes. 
For the Sergio, Greenhouse and the race car tasks, \textsc{SimSVGD[Only sim prior]} performs on par with \alg, which indicates that the prior over $\vphi$ is rich enough for these problems. However, in the Pendulum environment, 
we observe a gap in the performance of the two methods in high-data regimes. This illustrates the importance of having a GP to model the gap between the models. 
% \begin{figure}[ht!]
%     \centering
%     \includegraphics[width=\linewidth]{figures/regression_exp_low_to_high.pdf}
%     \caption{\looseness-1 On all the systems \alg achieves low Negative Log-Likelihood already in the low data regime and performs on par with the best from the rest also in high data regime. In Sergio, Greenhouse and Racecar environments, \textsc{SimFSVGD[Only sim prior]} also achieves the performance of \alg, which indicates that the prior over $\vphi$ is rich enough so that it can compensate the missing additive GP term. That does not happen in the Pendulum environment, showcasing the importance of having a GP to model the gap between the models.}
%     \label{fig:regression_low_to_high}
% \end{figure}

\paragraph{RC Car System Identification on Hardware}
Collecting data for learning dynamical systems in the real world is expensive. We investigate how \alg speeds up dynamics learning on a higly dynamics radio controlled (RC) car. 
The RC car is similar to the one in~\cite{sukhija2023gradient, bhardwaj2023data} and has a six-dimensional state (position, orientation, and velocities) and two-dimensional input (steering and throttle). We control the car at 30Hz. and use the Optitrack for robotics motion capture system
\footnote{\href{https://optitrack.com/applications/robotics/}{https://optitrack.com/applications/robotics/}}
to estimate the state. 
The car has a significant delay (ca. 80ms)
between transmission and execution of the control signals. 
Therefore, we include the last three actions $[\va_{t-3}, \va_{t-2}, \va_{t-1}]$ in the current state $\vs_t$. The resulting state space is $12$ dimensional and the action space is $2$ dimensional.
We report the final result in \cref{fig:panel_regression}. We observe that \alg outperforms all baselines in the experiments, and captures the highly nonlinear dynamics of the car with much fewer data points. This illustrates the sample-efficiency gains of \alg over state-of-the-art Bayesian dynamics learning methods.
% \begin{figure}[ht!]
%     \centering
%     \includegraphics[width=\linewidth]{figures/regression_exp_low_to_real.pdf}
%     \caption{\looseness-1 Also on the hardware dynamics, \alg achieves low Negative Log-Likelihood score already in low data regime and performs on par with best of the rest in the high data regime. \alg performs the best also with respect to the Root MSE score, with only \textsc{SimSVGD[Only sim prior]} matching its performance in the high data regime.}
%     \label{fig:regression_low_to_real}
% \end{figure}
\subsection*{Does \alg improve Sample Efficiency in Reinforcement Learning?}
\label{subsection: Application to Reinforcement Learning}
\begin{figure}[ht!]
    \centering
      \begin{overpic}[width=\textwidth]{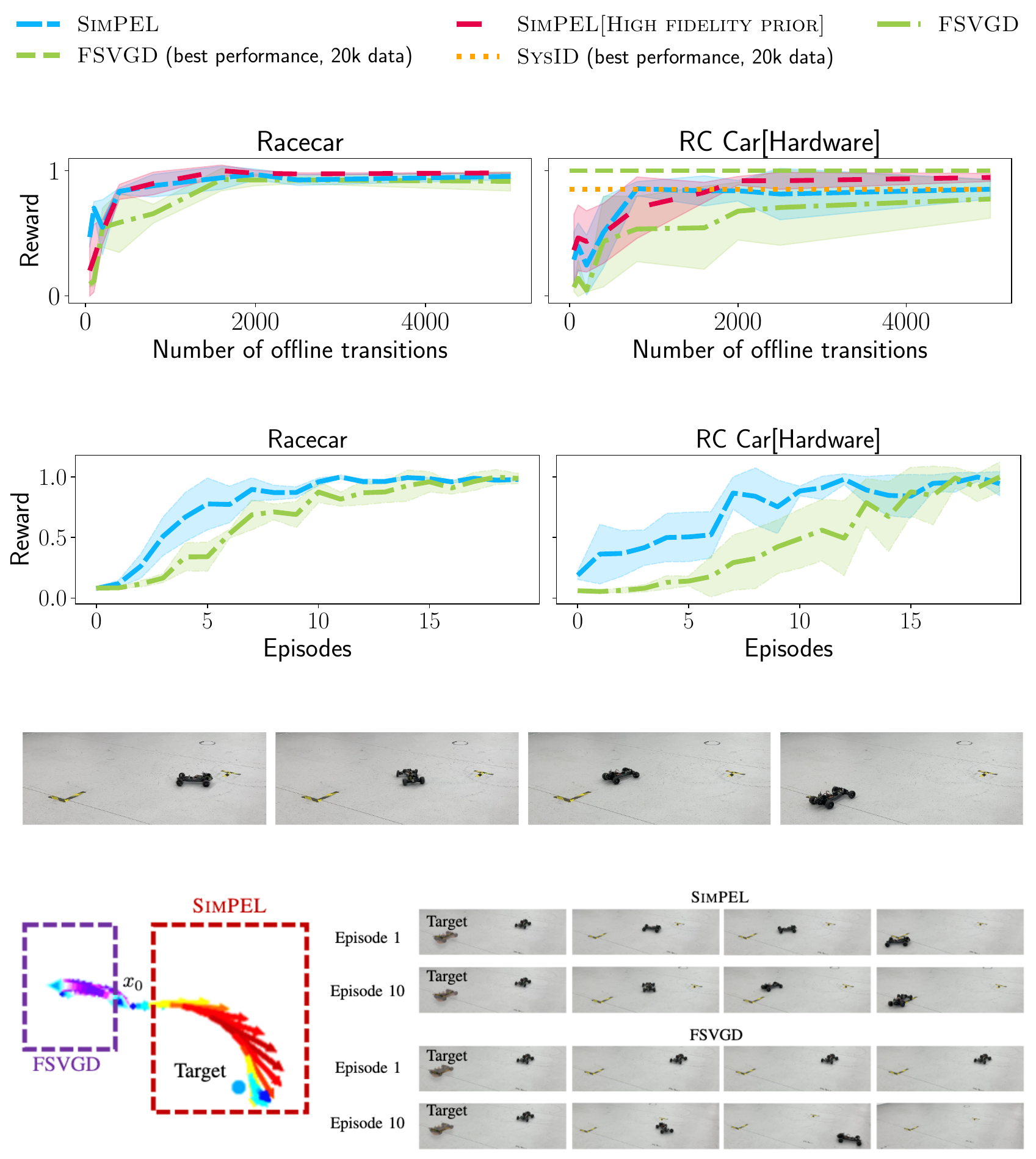}
    \put(0,90){\small \textbf{a) Offline RL}}
    \put(0,68){\line(1,0){90}}
    \put(0,64){\small \textbf{b) Online RL}}
    \put(0,42){\line(1,0){90}}
    \put(0,39){\small \textbf{c) Desired reverse parking maneuver (rotating the car $180^{\circ}$ and parking ca.~ 2m away)}}
    \put(0,27.5){\line(1,0){90}}
    \put(0,24.5){\small \textbf{d) Racecar Episode 3}}
    \put(28,24.5){\small \textbf{e) Realized trajectories of RC Car}}
    \put(29,11.8){\line(1,0){61}}

  \end{overpic}
    \caption{\looseness-1 
    (a) We train a dynamics model using an offline dataset and train a feedback NN with the model. \alg and \textsc{SimPEL[High fidelity prior]} achieve high rewards in both the low and high data regimes. %small data regime, while at the convergence performance of all algorithms that use NN dynamics model is the same. 
    %On RC Car hardware we see the sim-to-real gap between the policies trained using NN dynamics and dynamics obtained by fitting parameters of a high-fidelity model.
    (b) The simulation prior helps initially by exploring more directly towards the target. Both in simulation and on hardware the agent is more sample-efficient.
    (c) Desired reverse parking maneuver.
    (d) \alg approaches the target within three episodes, while FSVGD (no prior) explores away from it.
    (e) \alg nears perfect parking by episode 10; FSVGD fails to initiate proper movement and continues to diverge.
    }
    \label{fig:panel_rl}
\end{figure}
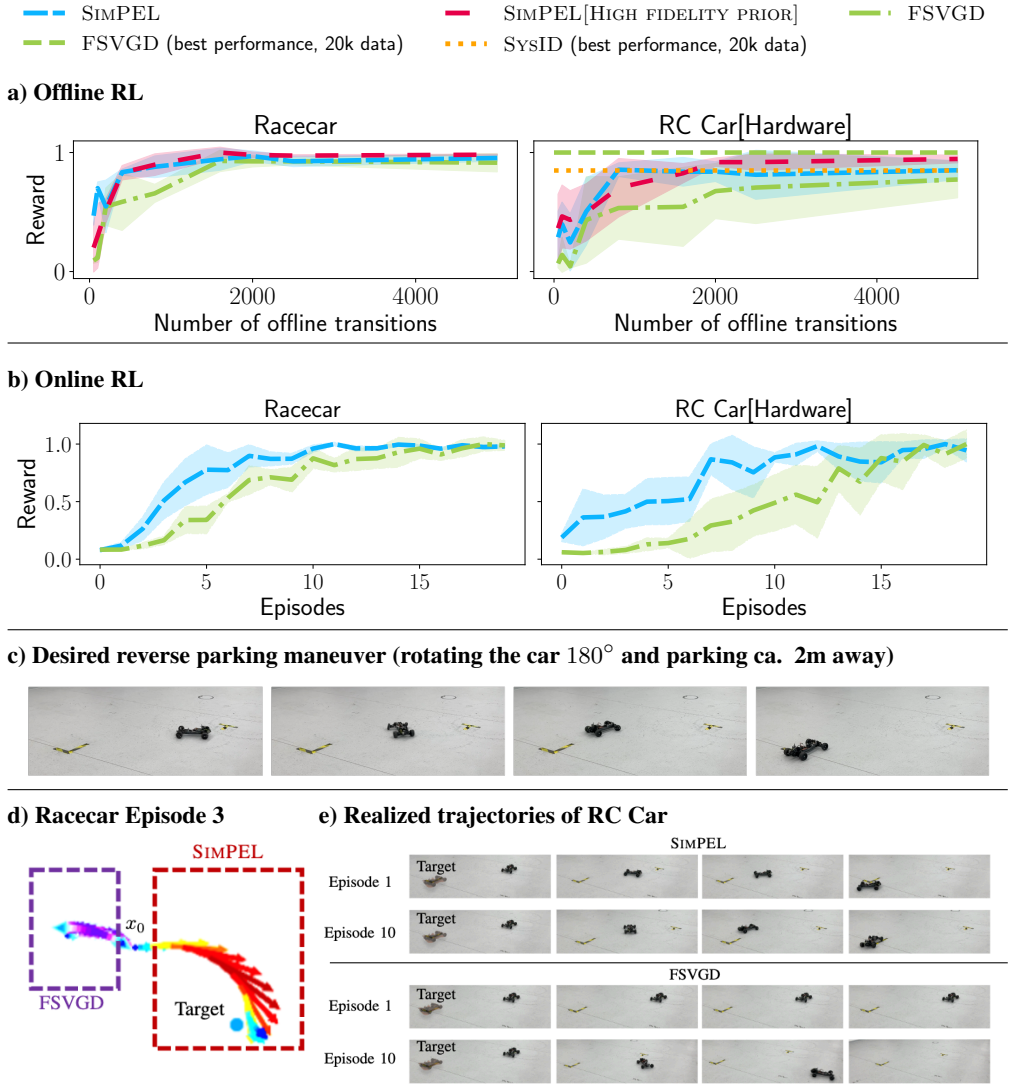

Finally, we apply \alg to learn the dynamics in the model-based reinforcement learning setting. We want to solve a reverse parking maneuver as depicted in \Cref{fig:panel_rl} (c). For the dynamics prior we use a simple kinematics bicycle model from \citet{kabzan2020amz}. We evaluate our method on the nonlinear dynamics model from~\citet{kabzan2020amz}, which also captures tyre dynamics with the Pacejka tyre model~\cite{pacejka1992magic} and the RC car hardware. To solve the reverse parking maneuver, we apply a sequence of 100 actions through a closed-loop neural network policy that we learn using SAC \citep{haarnoja2018soft} trained on the learned dynamics.

% \begin{figure}[ht!]
%     \centering
%     \includegraphics[width=\textwidth]{figures/parking_perfect.png}
%     \caption{Desired reverse parking maneuver which involves rotating the car $180^{\circ}$ and parking ca.~ 2m away}
%     \label{fig:parking_demo}
% \end{figure}

\paragraph{Offline Reinforcement Learning}
\label{subsec:offlinerl}
In the offline reinforcement learning setting, we first manually collect transitions and then fit a BNN model to obtain a posterior, conditioning on all the observed data. We use the learned posterior to train a control policy using SAC~\citep{haarnoja2018soft}. For generating rollouts from our model, we follow the trajectory sampling approach from~\citet{chua2018deep}. Finally, we evaluate the performance of the learned feedback policy on the system.

The results are presented in \Cref{fig:panel_rl}. On the left, we test the algorithms in simulation. Besides \alg and \textsc{FSVGD}, we also evaluate \textsc{SimFSVGD[High fidelity prior]}, a version of \alg that uses the high fidelity model as prior. 
From the experiments, we conclude that integrating the prior helps significantly in the low-data regime. This is particularly observed in our hardware experiments (right side of \Cref{fig:panel_rl}), where \alg obtains the highest performance across all baselines. Moreover, \textsc{SimFSVGD[High fidelity prior]} also outperforms the \textsc{SysID} baseline and matches the performance of \textsc{FSVGD} which is trained with $5 \times$ more data.

% \begin{figure}[ht!]
%     \centering
%     \includegraphics[width=\linewidth]{figures/offline_rl.pdf}
%     \caption{\looseness-1 Offline RL on high-fidelity Racecar simulator and hardware RC Car. We collect offline data manually, train a dynamics model, and use it to obtain feedback NN policy. Algorithms \alg and \textsc{SimFSVGD[High fidelity prior]}, that use simulator prior, achieve high reward on the environment already in a small data regime, while at the convergence performance of all algorithms that use NN dynamics model is the same. On RC Car hardware we see the sim-to-real gap between the policies trained using NN dynamics and dynamics obtained by fitting parameters of a high-fidelity model.}
%     \label{fig:offline_rl}
% \end{figure}

\paragraph{Episodic Model-Based Reinforcement Learning}
\looseness=-1
In the episodic model-based reinforcement learning setting, we learn online without any offline collected data.
In each episode, we first learn a dynamics model from the collected data thus far and then learn a policy using the learned dynamics.
We use a BNN for dynamics learning, where the prior in the case of \alg is the bicycle model of \citet{kabzan2020amz} with additive Gaussian process to cover for the sim-to-real gap. We compare our method to \textsc{FSVGD}. We use SAC with model generated rollouts for training the control policy.
%We learn the SAC \citep{haarnoja2018soft} policy, where for sampling the transitions we use the PETS \citep{chua2018deep} strategy using the trained BNN dynamics. 
We rollout the policy on the real system for 100 steps and add the data collected from rollouts to a data buffer.

% \begin{figure}[ht!]
%     \centering
%     \includegraphics[width=\linewidth]{figures/online_rl.pdf}
%     \caption{\looseness-1 The simulation prior helps initially by exploring more directly towards the target. Both in simulation and on hardware the agent solves the task significantly faster.}
%     \label{fig:online_rl}
% \end{figure}

In \Cref{fig:panel_rl}, we report the learning curves for the race car in simulation and the RC car hardware. We observe that in both cases, incorporating the simulation prior leads to considerably faster convergence (ca.~$2\times$ fewer episodes/hardware experiments).
Moreover, in \Cref{fig:panel_rl} (d) we show the realized trajectory for the third episode of the race car. We see that already after three episodes, \alg learns to drive the car close to the target whereas not incorporating the prior leads to exploration away from the target position. Similar behavior is observed on hardware (cf.,~\Cref{fig:panel_rl} (e)), where from episode 1, the policy learned using \alg drives the car close to the target, and at episode 10 it is parking the car nearly perfectly. On the contrary, not incorporating the prior leads to no movement of the car in episode 1 and the car driving away from the target in episode 10. This experiment demonstrates that incorporating a simulation prior in model-based RL leads to directed exploration which results in better sample-efficiency.

% \begin{figure*}[ht!]
%    \begin{center}
%     \includegraphics[width=\linewidth]{figures/panel_1.pdf}
% \end{center}
% \caption{
% (a) Realized trajectory in the race car simulation environment after episode 3. We observe that \alg already learns to approach the target within just three episodes, while FSVGD (without a prior) continues to explore away from it.  
% (b) Realized trajectories on hardware after episodes 1 and 10 for both \alg and FSVGD. \alg moves toward the target in the first episode and parks nearly perfectly by episode 10. In contrast, FSVGD fails to initiate movement in the first episode and continues to explore away from the target by episode 10. Full trajectories can be viewed in the video provided in the supplementary materials.
% }
% \label{fig:hw_online_rl}
% \vspace{-0.2cm}
% \end{figure*}

\section*{Methods}
%\label{section: Methods}
%%
% \section*{Background: Bayesian Neural Networks}
% \label{sec:background_fbnns}
\subsection*{Bayesian Deep Learning}
% We focus on a general regression problem with a dataset $\mathcal{D}=(\mathbf{X}^{\mathcal{D}},\mathbf{y}^{\mathcal{D}})$, comprised of $m$ noisy evaluations $\by_{j}=\bh^*\left(\bx_{j}\right) + \vepsilon_j$ of an unknown function $\bh^*: \calX \mapsto \calY$. The training inputs are denoted by $\mathbf{X}^{\mathcal{D}} = \left\{ \bx_j \right\}_{j=1}^m$ and corresponding function values by $\mathbf{y}^{\mathcal{D}} = \left\{ \by_j \right\}_{j=1}^m$. 
% The noise $\set{\vepsilon_j}_{j=1}^m$is assumed to be i.i.d.

% We employ a NN model $\bh_\vtheta: \calX \rightarrow \calY$ with weights $\vtheta \in \Theta$. 

We model the regression functions $\bh$ with BNNs. The conditional predictive distribution for the noisy observations $p(\vy|\bx,\vtheta)$ can be modeled with any likelihood function, however we model it with Gaussian, i.e., $p(\vy|\bx,\vtheta) = \calN(\vy|\bh_\vtheta(\bx), \sigma^2\mI)$, where $\sigma^2$ is the observation noise variance. 
The goal is to learn the parameters $\vtheta$ that represent the true function $\bh^*$ well. To this end, a common approach is to select $\vtheta$ that maximizes the data likelihood, i.e.,
\begin{equation}
    \vtheta^{\text{MLE}} = \argmax_{\vtheta \in \Theta} p(\mathbf{y}^{\mathcal{D}}|\mathbf{X}^{\mathcal{D}},\vtheta) \label{eq: mle theta}.
\end{equation}
Alternatively, given a prior distribution $p(\vtheta)$, we can select the $\vtheta$ that maximizes the posterior distribution,
\begin{equation}
    \vtheta^{\text{MAP}} = \argmax_{\vtheta \in \Theta} p(\vtheta| \mathcal{D}) = \argmax_{\vtheta \in \vtheta} p(\mathbf{y}^{\mathcal{D}}|\mathbf{X}^{\mathcal{D}},\vtheta)p(\vtheta) \label{eq: map theta}.
\end{equation}
The prior $p(\vtheta)$ is often used as a regularization to avoid overfitting the training data. A common choice for the prior is  $p(\vtheta) \sim \calN(0, \lambda^2 \bI)$, effectively resulting in a $l_2$-regularization of the parameter weights.

However, both \cref{eq: mle theta} and \cref{eq: map theta} result in a point estimate for the parameter $\vtheta$ and do not maintain a distribution over our belief of $\vtheta$. Hence, they do not capture our confidence in the proposed parameters. 
BNNs are used to maintain a distribution over the NN parameters $\vtheta$. In particular, Bayes' theorem combines the prior distribution $p(\vtheta)$ with the empirical data into a posterior distribution $p(\vtheta|\mathbf{X}^{\mathcal{D}}, \mathbf{y}^{\mathcal{D}}) \propto p(\mathbf{y}^{\mathcal{D}}|\mathbf{X}^{\mathcal{D}},\vtheta) p(\vtheta)$. %Here, $p(\mathbf{y}^{\mathcal{D}}|\mathbf{X}^{\mathcal{D}},\vtheta)$ is the likelihood of the data, given a NN hypothesis $\vtheta$. Under the i.i.d.~hypothesis, the likelihood factorizes as $p(\mathbf{y}^{\mathcal{D}}|\mathbf{X}^{\mathcal{D}},\vtheta) = \prod_{j=1}^m p(y_j|\bx_j,\vtheta)$.

To make predictions for an unseen test point $\bx^*$, we marginalize out the parameters $\vtheta$ in the posterior, i.e., the predictive distribution is calculated as 
\begin{align*}
    p(\vy^*|\bx^*,\mathcal{D}) &= \int p(\vy^*|\bx^*,\vtheta) p(\vtheta|\mathcal{D}) d\vtheta \\
    &= \E_{\vtheta}\left[ p(\vy^*|\bx^*,\vtheta) | \mathcal{D} \right].
\end{align*}

\subsection*{BNN Inference in Function Space}
The parameter space $\Theta$ is generally very high-dimensional and NNs are often over-parameterized, which makes posterior inference for BNNs extremely challenging.
Hence, {\em approximate inference} \citep{Blei2016, chen2018unified} techniques are devised to make BNN inference tractable. However, a crucial challenge with BNNs is that specifying a meaningful prior $p(\vtheta)$ is very hard and unintuitive \citep{noci2021precise}. This is firstly due to the high dimensionality and over-parameterization discussed above but also because mostly we only have a prior for the function output itself, e.g., smoothness,  periodicity, linearity, etc.
%Moreover, the complex parametrization also makes choosing an appropriate prior distribution $p(\vtheta)$ very hard.

Aiming to alleviate these issues, an alternative approach views BNN inference in the function space, i.e., the space of regression functions $\bh: \calX \mapsto \calY$ rather than in the parameter space $\Theta$. This results in the posterior $p(\bh |\mathcal{D}) \propto p(\mathbf{y}^{\mathcal{D}} | \mathbf{X}^{\mathcal{D}},\bh) p(\bh)$ \citep{wang2019function, sun2019functional}. Here, $p(\bh)$ is a {\em stochastic process} prior, e.g., Gaussian process (GP)~\citep{rasmussen2003gaussian}. 

\looseness -1 Stochastic processes can be viewed as infinite-dimensional random vectors, making direct representation of \( p(\vh) \) intractable. 
However, they are fully characterized by their finite-dimensional marginal distributions, as formalized by the Kolmogorov Extension Theorem \citep{stochastic_differential_equations}. 
For any \emph{measurement set} $\mathbf{X} := [\bx_1, \dots, \bx_m] \in \calX^m $, consisting of $ m \in \mathbb{N} $ points in the input space $ \calX $, the process is specified by the joint distribution over the corresponding function values, $p(\mathbf{h}^\mathbf{X}) := p(h(\bx_1), \dots, h(\bx_m))$. 
When the dataset inputs $\mathbf{X}^{\mathcal{D}}$ are part of the measurement set, i.e. $\mathbf{X}^{\mathcal{D}} \subset \mathbf{X}$, this perspective enables a more tractable approach to functional BNN inference by reformulating it in terms of posterior marginals over measurement sets $\bX$: $p(\rvh^\bX \mid \mathbf{X}, \mathcal{D}) \propto p(\mathbf{y}^{\mathcal{D}} \mid \rvh^{\bX^{\mathcal{D}}}) \, p(\rvh^{\bX}).$
While our focus is on BNNs, the expression $ p(\mathbf{y}^{\mathcal{D}} \mid \rvh^{\bX^{\mathcal{D}}}) \, p(\rvh^{\bX}) $ can also be used to derive a MAP estimate in function space, analogous to \cref{eq: map theta}, by specifying an appropriate functional prior.

\looseness -1 To facilitate functional BNN inference, the functional posterior $p(\rvh^\bX |\mathbf{X}, \mathcal{D})$ can be tractably approximated by Bayesian inference techniques such as~\citep{sun2019functional, wang2019function}. In this work, we use the functional Stein Variational Gradient Descent \citep[\textsc{FSVGD},][]{wang2019function} method due its simplicity and good uncertainty quantification. However, the \alg framework is independent to the choice of inference algorithm.
The \textsc{FSVGD} method approximates the posterior as a set of $L$ NN parameter particles $\{ \vtheta_1, \dots, \vtheta_L \}$. 
To improve the particle approximation, \textsc{FSVGD} first iteratively re-samples measurement sets $\widehat{\bX}$ from a measurement distribution $\bm{\zeta}$, supported on $\calX$, e.g. $\operatorname{Unif}(\calX)$, 
then joins the sampled measurement set with the  dataset inputs $\bX = [\mathbf{X}^{\mathcal{D}}, \widehat{\bX}]$,
and updates the parameters using the following update rule:
\begin{equation} \label{eq:fsvgd_updates}
    \vtheta^i \leftarrow \vtheta^i + \gamma  \underbrace{\Big( \nabla_{\vtheta^i} \rvh_{\vtheta^i}^\bX \Big)^\top}_{\text{NN Jacobian}}  \bigg( \underbrace{\frac{1}{L} \sum_{l=1}^L \bK_{li} \nabla_{\rvh_{\vtheta^l}^\bX} \ln p(\rvh_{\vtheta^l}^\bX |\mathbf{X}, \mathcal{D}) + \nabla_{\rvh_{\vtheta^l}^\bX} \bK_{li}}_{\text{SVGD update in the function space} \vspace{-4pt}} \bigg) ~ ,
\end{equation}
where $\gamma$ is the learning rate, $\nabla_{\rvh^\bX} \ln p(\rvh^\bX |\mathbf{X},  \mathcal{D})$ is the functional posterior score, i.e., the gradient of the log-density
\begin{equation}
\label{eq: gradient log density}
    \nabla_{\rvh^\bX} \ln p(\rvh^\bX |\mathbf{X},  \mathcal{D})  = \nabla_{\rvh^\bX} \ln p(\mathbf{y}^{\mathcal{D}} | \rvh^{\bX^{\mathcal{D}}}) + \nabla_{\rvh^\bX} \ln p(\rvh^{\bX}) ~,
\end{equation} 
and $\mathbf{K} = [k(\rvh_{\vtheta^l}^\bX, \rvh_{\vtheta^i}^\bX)]_{li}$ is the kernel matrix between the function values in the measurement points based on a kernel function $k(\cdot, \cdot)$. 
The gradient $\nabla_{\rvh^\bX} \ln p(\mathbf{y}^{\mathcal{D}} | \rvh^{\bX^{\mathcal{D}}})$ term in \Cref{eq: gradient log density} is of the form $[\nabla_{\rvh^{\bX^\calD}} \ln p(\mathbf{y}^{\mathcal{D}} | \rvh^{\bX^{\mathcal{D}}}), \bf{0}]$.
Effectively, the update rule in \cref{eq:fsvgd_updates} weights gradients of parameters w.r.t.~their closeness in the function space as measured by $k$, i.e., the term: $ \bK_{li} \nabla_{\rvh_{\vtheta^l}^\bX} \ln p(\rvh_{\vtheta^l}^\bX |\mathbf{X}, \mathcal{D})$ while also avoiding mode collapse of the particles by encouraging diversity, the term: $\nabla_{\rvh_{\vtheta^l}^\bX} \bK_{li}$. 

Given a point $\bx^*$, we approximate $\E_{\vh}[\vy^*|\bx^*,\mathcal{D}]$ and  $\text{Var}_{\vh}[\vy^*|\bx^*,\mathcal{D}]$ as 
\begin{align}
    \E_{\vh}[\vy^*|\bx^*,\mathcal{D}] \approx \frac{1}{L}\sum_{i=1}^L \bh_{\vtheta_i}(\bx^*), \quad
    \text{Var}_{\vh}[\vy^*|\bx^*,\mathcal{D}] \approx \text{Var}\left(\{\bh_{\vtheta_i}(\bx^*)\}^L_{i=1}\right)
\end{align}

Crucially \textsc{FSVGD} only uses the prior scores $\nabla_{\rvh^\bX} \ln p(\rvh^\bX )$ to update the model and does not require the density function of the prior marginals. This is also the case for other approximate functional inference techniques and constitutes a key insight that we will later draw upon in our approach.

% \subsection{Score Estimation}

% The goal of score estimation is to approximate the  score $\bs_p(\bx) = \nabla_\bx \log p(\bx)$ of an unknown probability distribution $p(\bx)$ based on i.i.d samples $\bX = [\bx_1, ..., \bx_l, ..., \bx_m],~ \bx_l \sim p(\bx)$ from the distribution.

% In an idealized case where we have the ground truth values of $\bs_p$ at the sample locations, we could estimate the score via kernelized, vector-valued regression:
% %
% \begin{equation}
%     \hat{\bs}_p = \argmin_{\bs \in \calH_\calK} \frac{1}{m} \sum_{j=1}^m \norm{s(\bx_j) - s_p(\bx_j)}^2
% \end{equation}

\subsection*{Estimating the Stochastic Process Prior Score}
\looseness=-1
\label{subsection: Estimating the Stochastic Process Prior Score}
Analytically computing the score for the stochastic prior is intractable for most domain-model processes. To this end, we take a sampling-based approach for score estimation. We draw $N$ samples $\rvh_{i, j}^{\bX} \sim \rvh_i^{\bX}$ for $j \in \set{1, \ldots, N}$ and use them to estimate the score $\nabla_{\rvh_i^\bX}\ln p(\rvh_i^{\bX})$.
We discuss both parametric and nonparametric approaches for score estimation~\citep[e.g.][]{shi2018spectral, zhou2020nonparametric} below.

\paragraph{Approximation with Gaussian}
A very simple and computationally cheap approach to estimate the score $\nabla_{\rvh_i^\bX}\ln p(\rvh_i^{\bX})$ is to approximate the distribution $p(\rvh_i^{\bX})$ with a Gaussian, i.e.
    \begin{align*}
        p(\rvh_i^{\bX}) &\approx \normaldist{\rvh_i^{\bX}}{\vmu^{\bX}_i}{\mSigma^X_i}, \\
        &\vmu^{\bX}_i = \frac{1}{N}\sum_{j=1}^N\rvh_{i, j}^{\bX}, \quad \mSigma^X_i = \frac{1}{N-1}\sum_{j=1}^N\left(\rvh_{i, j}^{\bX}- \vmu_i^\bX\right)\left(\rvh_{i, j}^{\bX}- \vmu_i^\bX\right)^\top
    \end{align*}
The approximate score is the gradient of the normal distribution $\nabla_{\rvh_i^\bX} \normaldist{\rvh_i^{\bX}}{\vmu^{\bX}_i}{\mSigma^X_i}$ for which we have closed-form expressions.

\paragraph{Approximation with Kernel Density Estimation (KDE)}
Approximating the distribution with a Gaussian means that our estimate will be unimodal.
One way to model potentially multi-modal distributions is with kernel density estimation. Here we approximate the distribution $p(\rvh_i^{\bX})$ with 
\begin{align*}
    p(\rvh_i^{\bX}) \approx &\frac{1}{N}\sum_{j=1}^N\normaldist{\rvh_i^{\bX}}{\rvh_{i, j}^{\bX}}{\gamma^2 \mI},
\end{align*}
where $\gamma$ is the KDE bandwidth. Same as with the Gaussian approximation, we have the closed-form formulas for the  $\nabla_{\rvh_i^\bX} \frac{1}{N}\sum_{j=1}^N\normaldist{\rvh_i^{\bX}}{\rvh_{i, j}^{\bX}}{\gamma^2 \mI}$ for the computation of the approximate score.

\paragraph{Nonparametric Score Approximations}
Another approach is to estimate the score $\nabla_{\rvh_i^\bX}\ln p(\rvh_i^{\bX})$ nonparametrically, where we learn the score function $s_p(\rvh_i^\bX) = \nabla_{\rvh_i^\bX}\ln p(\rvh_i^{\bX})$.  Assume we have access to the score at our $N$ samples, i.e. we know $s_p(\rvh_{i, j}^\bX)$ for $j \in \set{1, \ldots, N}$. We can then estimate the score function with regression in the Reproducing Kernel Hilbert Space (RKHS) $\mathcal{H}_\mathcal{K}$, reproduced with the matrix kernel $\mathcal{K}$:
\begin{align}
\label{eq: rkhs regression}
    \widehat{s}_{p, \lambda} = \argmin_{s \in \mathcal{H}_\mathcal{K}} \frac{1}{N}\sum_{j=1}^N\norm{s(\rvh_{i, j}^\bX) - s_p(\rvh_{i, j}^\bX)}_2^2 + \frac{\lambda}{2} \norm{s}_{\mathcal{H}_\mathcal{K}}^2
\end{align}
 \Cref{eq: rkhs regression} has a closed-form solution; $\widehat{s}_{p, \lambda} = \left(\widehat{L}_{\mathcal{K}} + \lambda I\right)^{-1}\widehat{L}_{\mathcal{K}}s_p$, where the positive-semidefinite operator $\widehat{L}_\mathcal{K}$ is defined in such a way that for a function $f$ we have $\widehat{L}_{\mathcal{K}}f = \frac{1}{N}\sum_{j=1}^N \mathcal{K}(\rvh_{i, j}^\bX, \cdot)f(\rvh_{i, j}^\bX)$. In reality, we do not have access to the score values. However, under some mild regularity conditions,  \citet{zhou2020nonparametric} show that $\widehat{L}_{\mathcal{K}}s_p = -\frac{1}{N}\sum_{j=1}^N\divergence_{\rvh_{i, j}^\bX}\mathcal{K}(\rvh_{i, j}^\bX, \cdot)$. 
From the spectral decomposition of the semi-definite operator $\widehat{L}_{\mathcal{K}} = \bU \mathbf{\Sigma}\bU^*$, where $\bU$ is unitary operator and $\mathbf{\Sigma}$ diagonal with eigenvalues on the diagonal, we see that the term $\left(\widehat{L}_{\mathcal{K}} + \lambda I\right)^{-1}$ (called Tikhonov regularizer) changes the eigenvalues of the operator $\widehat{L}_{\mathcal{K}}$ with the function $g_{\lambda}(\sigma) = (\sigma + \lambda)^{-1}$, i.e., we have 
\begin{align}
\label{eq: Tikhonov regularizer}
\left(\widehat{L}_{\mathcal{K}} + \lambda I\right)^{-1} = \bU (\mathbf{\Sigma}+\lambda\bI)^{-1}\bU^* \overset{(\blacktriangle)}{=} \bU g_{\lambda}(\mathbf{\Sigma})\bU^* \defeq g_{\lambda}\left(\widehat{L}_{\mathcal{K}}\right).
\end{align} 
Here we applied function $g_{\lambda}$ elementwise to the diagonal operator $\mathbf{\Sigma}$ in $(\blacktriangle)$.
We can replace the Tikhonov regularization with some other function $g_{\lambda}$ by applying it elementwise to the diagonal operator $\mathbf{\Sigma}$ to approximate the inversion of the eigenvalues of the operator $\widehat{L}_{\mathcal{K}}$. This results in the estimator:
\begin{align*}
    \widehat{s}_{p, \lambda}^g = -\frac{1}{N}g_{\lambda}\left(\widehat{L}_{\mathcal{K}}\right)\sum_{j=1}^N\divergence_{\rvh_{i, j}^\bX}\mathcal{K}(\rvh_{i, j}^\bX, \cdot).
\end{align*}
Besides the design choice of the regularization function $g_{\lambda}$ we also have the design choice of the matrix kernel $\mathcal{K}$ that reproduces the space of functions $\mathcal{H}_{\mathcal{K}}$ over which we are optimizing the score estimator $\widehat{s}_{p, \lambda}^g$. In our experiments, we benchmark two nonparametric score estimation techniques, namely Spectral Stein Gradient Estimation (SSGE) \citep{shi2018spectral} and the $\nu$-method \citep{zhou2020nonparametric}.

The SSGE method applies the spectral cut-off regularization, i.e., $g_\lambda(\sigma) = \mathbf{1}_{\set{\sigma \ge \lambda}}\sigma^{-1}$ and uses the diagonal kernel $\mathcal{K}(\bx, \by) = k(\bx, \by)\mI$, where $k$ can be any scalar kernel (in our experiments we picked Gaussian kernel). 

To incorporate the fact that the score function $s_p$ is a gradient function, the $\nu$-method derives the matrix kernel $\mathcal{K}$ from a translation-invariant kernel $k(\bx, \by) = \phi(\bx-\by)$ (in our experiments a Gaussian kernel) as $\mathcal{K}(\bx, \by) = -\nabla^2\phi(\bx-\by)$.
With this design choice, every function in $\mathcal{H}_{\mathcal{K}}$ is a gradient of some function.
% i.e. $\forall \vf \in \mathcal{H}_{\mathcal{K}}, \exists \vg, \text{s.t.} \vf = \nabla \vg$.
% With this design choice, any function in $\mathcal{H}_{\mathcal{K}}$ is a gradient of some function.
The $\nu$-method computes $g_{\lambda}\left(\widehat{L}_{\mathcal{K}}\right)$ iteratively and the regularization function $g_{\lambda}(\sigma)$ that approximates $1/\sigma$ is represented with the family of polynomials, i.e., $g_{\lambda}(\sigma) = \poly(\sigma)$.

% \subsection*{The \alg algorithm}
% Finally, combining the stochastic process prior, derived from the simulation prior in \Cref{subsection: Incorporating simulators as functional priors} with the score estimation techniques from \Cref{subsection: Estimating the Stochastic Process Prior Score}, we arrive at the main algorithm.

% \begin{algorithm}[th]
% \caption{\strut \alg}\label{alg:sim_transfer_alg}
% \hspace*{\algorithmicindent} \textbf{Input:} Measurement distribution
% $\mu$, Simulation prior $\bg$, Parameter distribution $p(\vphi)$, GP 
% \hspace*{\algorithmicindent} \hspace{2.7em} $p(\tilde{\bh})$, Data $\calD$, BNN particles $\{\vtheta_i\}^{L}_{i=1}$
% \begin{algorithmic}[1]
% \State Sample measurement set $\bX = [\mathbf{X}^{\mathcal{D}}, \widehat{\bX}], \widehat{\bX}  \sim \bm{\zeta}$
% \State Sample simulation prior function values $\{\bh^{\bX}_{i, j}\}^{d_y}_{i=0}$ for $j \in \{1, \dots, N\}$.
% \State Approximate the prior score term $\nabla_{\rvh_{\vtheta^l}^\bX} \ln p(\rvh_{\vtheta^l}^\bX)$ for $l \in \set{1, \ldots, L}$ with samples from the prior $\{\bh^{\bX}_{i, j}\}^{d_y}_{i=0}$ for $j \in \{1, \dots, N\}$ using techniques from \Cref{subsection: Estimating the Stochastic Process Prior Score}.
% \State Update BNN particles with \cref{eq:fsvgd_updates}.
% \end{algorithmic}
% \end{algorithm}

\paragraph{Comparison of Score Estimation Techniques}

\begin{figure}[ht!]
    \centering
    \includegraphics[width=\linewidth]{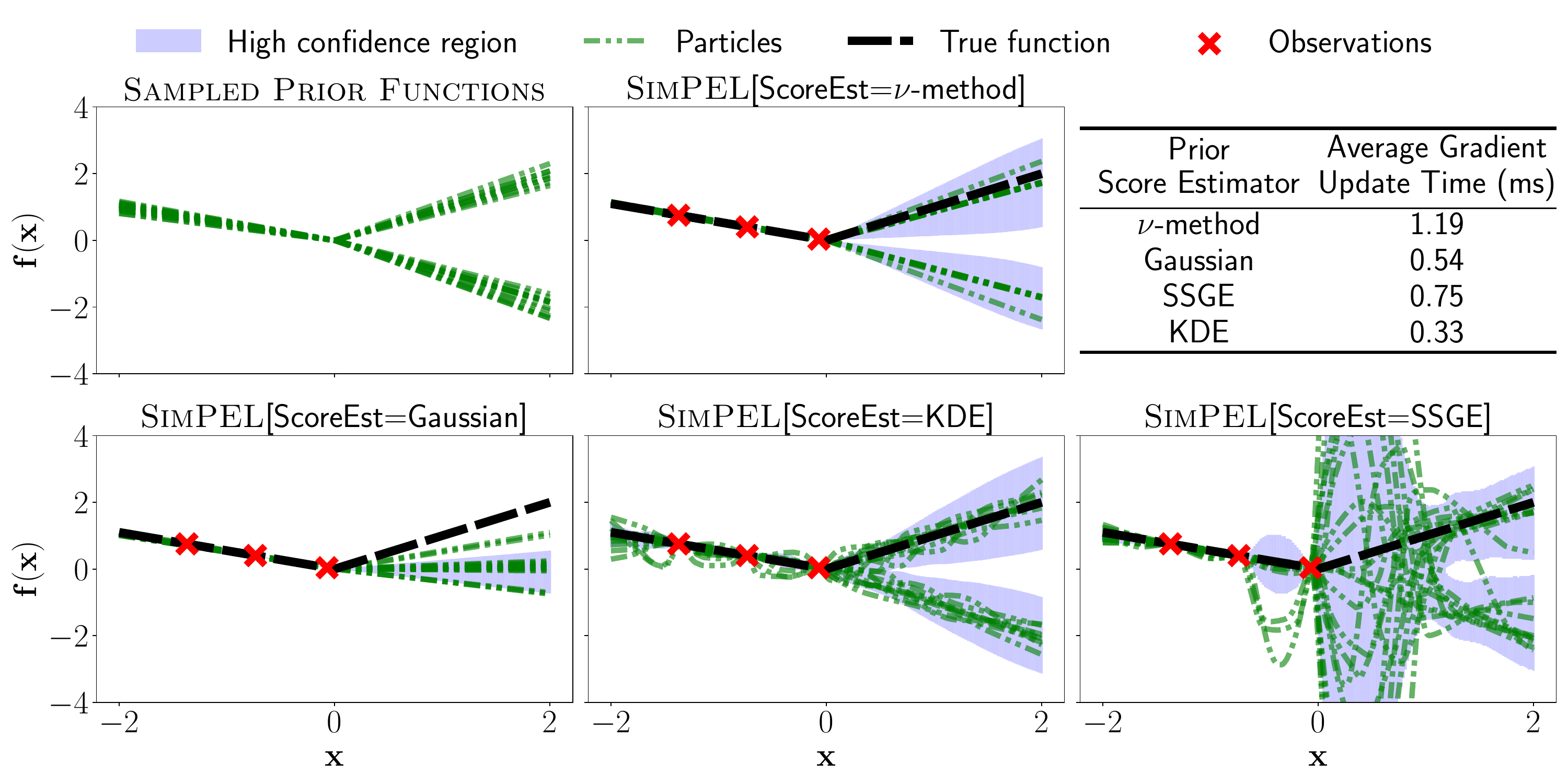}
    \caption{Illustration of bi-model functional prior.
    %where part of the function samples are multimodal. 
    While the $\nu$-method successfully models both posterior modes, the Gaussian score estimator can model only one mode. KDE and SSGE can also model several modes, however, their particle paths do not resemble the linear structure of the possible posterior function. The time for the gradient update step is the largest for the $\nu$-method and lasts approximately 4 times more than the gradient update step if we use KDE score estimator which is the fastest in our case. 
}
    \label{fig:multimodal_example}
\end{figure}

We illustrate the performance of \alg coupled with different prior score estimators on multi-modal prior example in \Cref{fig:multimodal_example}. As we see in the \Cref{fig:multimodal_example}, estimating the prior score with $\nu$-method can capture larger class of distributions than approximation with a Gaussian, however, in practice, already the Gaussian approximation results in satisfactory performance. Hence, in all our experiments we used the Gaussian approximation for prior score matching. If one knows certain characteristics of the prior, such us multi-modality, with which Gaussian approximation would underperform, then we advise using the $\nu$-method for prior score computation.

\section*{Conclusion}
In this work, we present \alg, an algorithm that incorporates low-fidelity simulators into the prior stochastic processes of BNN inference. 
To cover for the gap between the low-fidelity simulator and the true data-generating process, we additionally add a GP to the low-fidelity simulator prior. 
We demonstrate the efficiency of \alg in learning dynamics across a variety of applications, including robotics, single-cell and greenhouse dynamics. In addition to its use in system identification, we show that \alg effectively closes the sim-to-real gap on an RC car hardware, enabling efficient reinforcement learning for both offline and online settings. Notably, we obtain a $2\times$ reduction in the number of physical experiments required to solve a dynamic parking maneuver in the online reinforcement learning setting.
\newpage
% MAIN CONTENT END

% REFERENCES 
\bibliography{references}
%\bibliographystyle{style/iclr2024_conference}

% APPENDIX
% \appendix
% \input{content/appendix_dump}

\end{document}